\documentclass{article}
\usepackage{spconf,amsmath,graphicx}
\usepackage{amssymb}
\usepackage{multirow}
\usepackage{graphics}
\usepackage{adjustbox}
\usepackage{booktabs}
\usepackage{tabularx}
\usepackage{enumitem}
\usepackage{xcolor}
\usepackage{caption}
\usepackage{subcaption}
\usepackage{url}

\title{Pose-invariant Face Recognition via Feature-Space Pose Frontalization}

\name{Nikolay Stanishev$^*$, Yuhang Lu$^*$, and Touradj Ebrahimi\thanks{$^\ast$Equal contribution}\thanks{Support from the Swiss National Science Foundation (SNSF) 20CH21\_195532 for XAIface CHIST-ERA-19-XAI-011 is acknowledged.}}
\address{Multimedia Signal Processing Group (MMSPG)\\
\'Ecole Polytechnique F\'ed\'erale de Lausanne (EPFL)}

\begin{document}
%\ninept
%
\maketitle
\begin{abstract}
Pose-invariant face recognition has become a challenging problem for modern AI-based face recognition systems. It aims at matching a profile face captured in the wild with a frontal face registered in a database. Existing methods perform face frontalization via either generative models or learning a pose robust feature representation. In this paper, a new method is presented to perform face frontalization and recognition within the feature space. First, a novel feature space pose frontalization module (FSPFM) is proposed to transform profile images with arbitrary angles into frontal counterparts. Second, a new training paradigm is proposed to maximize the potential of FSPFM and boost its performance. The latter consists of a pre-training and an attention-guided fine-tuning stage. Moreover, extensive experiments have been conducted on five popular face recognition benchmarks. Results show that not only our method outperforms the state-of-the-art in the pose-invariant face recognition task but also maintains superior performance in other standard scenarios. 

\end{abstract}
\section{Introduction}
\label{sec:intro}

Recent breakthroughs in deep learning have significantly impacted the field of face recognition, leading to remarkable performance improvement on public benchmarks and in real-world applications. Despite these achievements, recent studies~\cite{lu2022novel} have shown a substantial degradation in the performance of current deep learning-based face recognition methods when face images are captured under varying poses. This challenge is known as the pose-invariant face recognition (PIFR) problem. In typical scenarios, face images are captured in a controlled environment with subjects facing directly toward the camera, which is referred to as ``frontal faces'' in this paper. In contrast, PIFR involves more complex scenarios where images are taken from various pose angles, defined as ``profile faces''.  

The challenge in PIFR arises from the substantial variations in facial appearance when images are captured from different angles. First, deep learning-based face recognition systems rely heavily on training data, which are often imbalanced in terms of pose distribution. This leads to a noticeable performance gap between frontal and profile face recognition. Second, as the facial pose shifts from frontal to profile, certain facial features become occluded, causing geometric distortions and spatial mismatching. As a result, the information extracted from frontal faces can differ substantially from that of profile faces, making the recognition task more challenging.

Several approaches have been proposed to address the PIFR problem. One primary line of research focuses on image-space face frontalization~\cite{zhu_high_2015, huang_beyond_2017, yin_towards_2017, zhao_towards_2018}. Such methods typically synthesize profile faces into frontal views to reduce pose variation before the recognition process. However, these methods tend to underperform when confronted with extreme profile poses. Moreover, they often introduce additional computational overhead due to the face transformation, laying a burden on the overall face recognition system.
An alternative approach seeks to learn a unified feature representation that remains robust across both frontal and profile face images~\cite{tran_disentangled_2017, cao_pose_2018, tsai_pam_2021, huang_attention_2021, mostofa2022pose, zhang_pose_2022}. Some have explored two-branch networks to process frontal and profile images respectively and jointly optimise them to mitigate pose-related discrepancies~\cite{tran_disentangled_2017, zhang_pose_2022}. Cao et al.~\cite{cao_pose_2018} first introduced a residual learning block that performs pose frontalization in the feature space. Several follow-up research~\cite{huang_attention_2021, tsai_pam_2021, mostofa2022pose} explored different designs of the specific feature transformation block, but they overlooked potential benefits of more effectively leveraging pose information and refining optimization strategies. 

To this end, this work introduces a novel approach to address the PIFR problem. First, a new feature transformation module FSPFM is proposed, which leverages pre-captured pose information as priors and automatically transforms profile faces with arbitrary angles into frontals  within the feature space. 
Moreover, a new two-stage training paradigm is conceived to maximize the effectiveness of the FSPFM. The first stage involves a pre-training process to learn general face representations, followed by a fine-tuning process with pair-wise supervision, which enhances FSPFM's ability to handle pose variations. To further refine the fine-tuning process, an attention-guided feature adaptation mechanism is proposed. This attention module re-weights the frontal features in a self-adaptive manner, allowing FSPFM to more precisely learn the residuals between profile and frontal features.

\begin{figure}{}
     \centering
     \begin{subfigure}[b]{\linewidth}
         \centering
         \includegraphics[width=\linewidth]{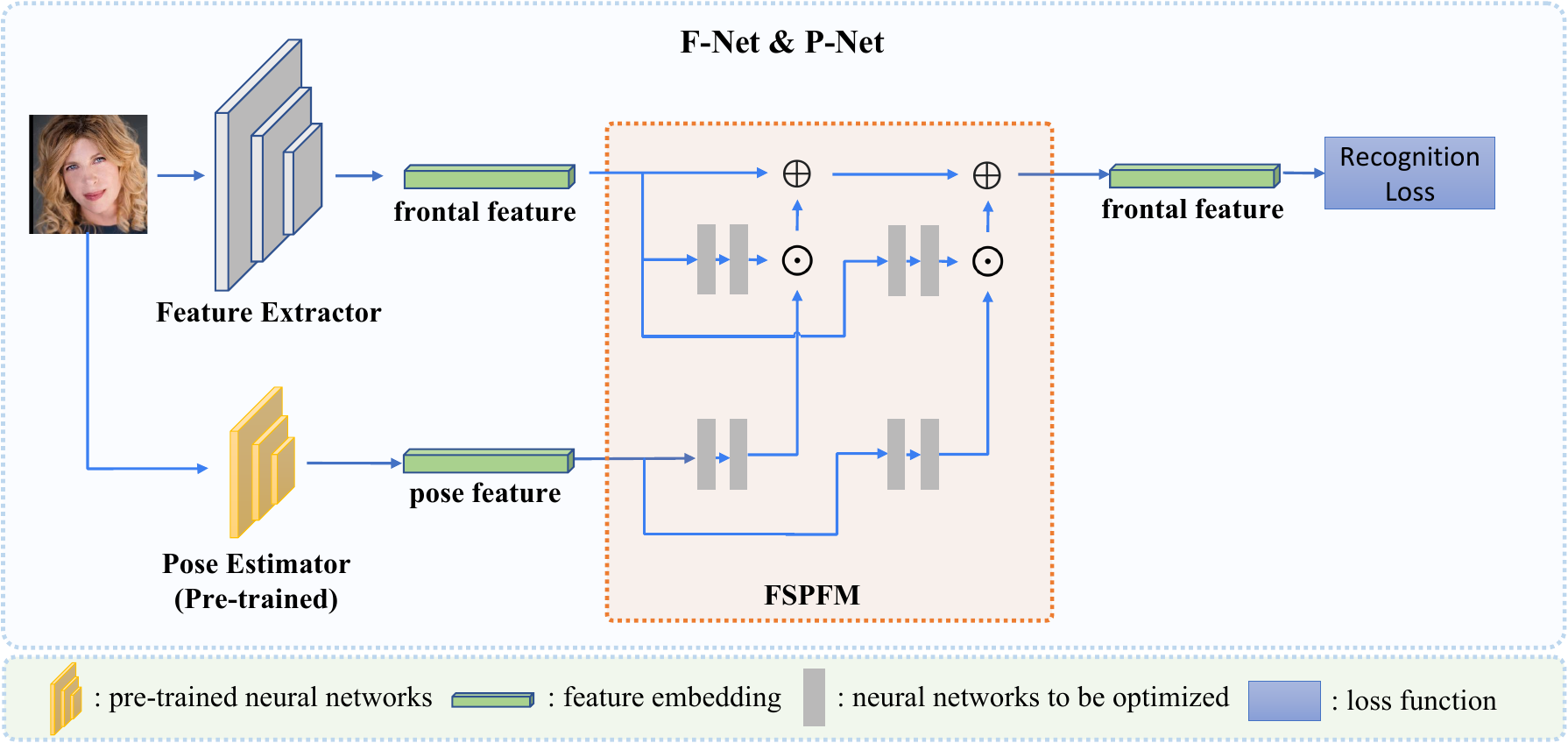}
         \caption{Stage 1: Pre-training the face feature extractor and pose frontalization module. }
         \label{fig:pretrain}
     \end{subfigure}
     \hfill
     \begin{subfigure}[b]{\linewidth}
         \centering
         \includegraphics[width=\linewidth]{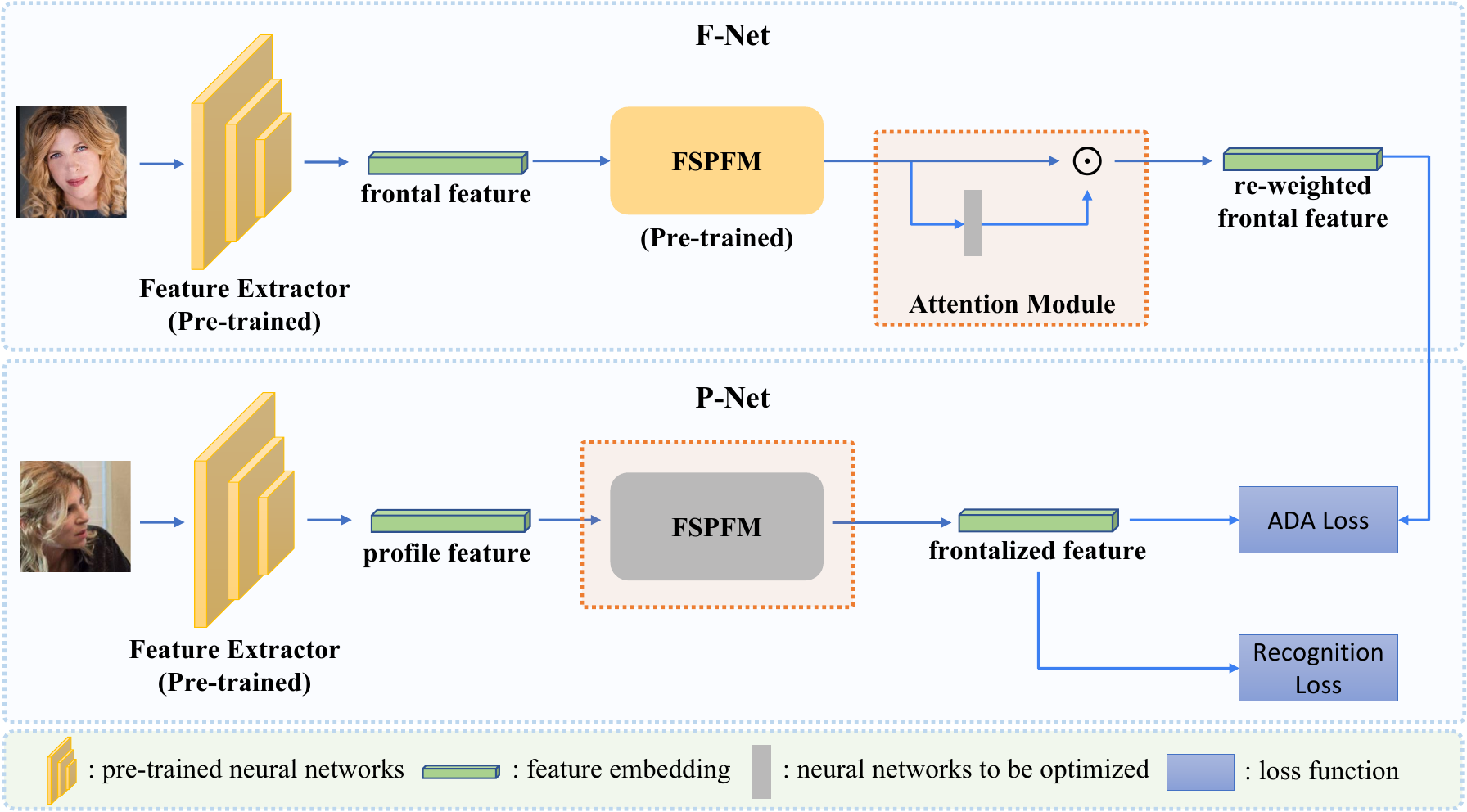}
         \caption{Stage 2: Fine-tuning the pose frontalization module in P-Net.}
         \label{fig:finetune}
     \end{subfigure}
     \caption{Pipeline of the proposed PIFR method. It consists of two stages of optimization, namely pre-training and fine-tuning. Consequently, the P-Net is regarded as the target model to perform PIFR. }
     \label{fig:pifr-pipeline}
\end{figure}

\section{Related Work}
\subsection{Face frontalization-based method}
Face frontalization refers to the process of synthesizing a frontal face image from arbitrary head poses. Earlier attempts mainly relied on computer graphics techniques. One typical solution~\cite{zhu_high_2015} involves fitting faces to a 3D Morphable Model (3DMM)~\cite{blanz_morphable_2023} and then directly rotating the model to generate frontal views. 
Recent studies have leveraged generative models to synthesize frontal faces, framing it as an image-to-image translation task. For example, Huang et al.~\cite{huang_beyond_2017} proposed TP-GAN that comprised a global and a local pathway to construct photorealistic frontal faces. Yin et al.~\cite{yin_towards_2017} proposed FF-GAN that integrated 3DMM into a GAN framework and additionally optimized a recognition loss to preserve identity information. PIM~\cite{zhao_towards_2018} retained the dual-path structure of TP-GAN and further developed more effective adversarial training strategies to enhance the quality of synthesized frontal faces.

\subsection{Pose-invariant feature representation-based method}
This approach aims to learn a unified feature representation across both frontal and profile images. 
Tran et al.~\cite{tran_disentangled_2017} proposed DR-GAN with disentangled encoder-decoder architecture that can learn pose-invariant representation. Zhang et al.~\cite{zhang_pose_2022} enhanced the PIFR training process by leveraging a knowledge distillation framework. It distills the angular knowledge of frontal faces to the profile faces, compactly clustering cross-pose features from the same person. %from the teacher network to the student network
Cao et al.~\cite{cao_pose_2018} presented the idea of pose transformation at the feature level instead of image level and proposed the DREAM (Deep Residual EquivAriant Mapping) method. They employed a soft gate learning strategy to model the transformation between frontal and profile faces. Inspired by a similar idea, Tsai and Yeh~\cite{tsai_pam_2021} proposed a lightweight Pose Attention Module (PAM), which learns the residual between pose variations of given inputs. Huang et al.~\cite{huang_attention_2021} proposed an attention-guided progressive transformation method that gradually converts a profile face representation to a frontal pose. Mostofa et al.~\cite{mostofa2022pose} used pose as side information via an attention module to guide the network to learn pose-invariant features.

\section{Proposed Method}
This work addresses the challenges of PIFR with two main contributions: a novel feature-space pose transformation module and a new training paradigm designed to maximize the potential of the proposed module.

\subsection{Feature space pose frontalization module (FSPFM)}
This subsection introduces FSPFM as our first original contribution to the PIFR problem. Figure~\ref{fig:pretrain} depicts an overview of the proposed module. This module first employs a pre-trained deep learning-based pose estimator to extract head pose information from input images. Then, it fuses the prior pose information with the extracted face features to perform frontalization in feature space. The key components are elaborated as follows.

\textbf{Head pose estimation:} 
To begin with, the head poses of input face images are estimated to serve as prior information for future transformation. The pre-trained HopeNet~\cite{ruiz_fine_2018} is employed to predict intrinsic Euler angles of face images that represent their 3D rotations. In principle, the HopeNet will first produce a universal pose feature representation, which can be disentangled into three coefficients representing yaw, pitch, and roll angles, respectively. Previous explorations~\cite{cao_pose_2018, huang_attention_2021} solely leveraged the yaw angle to serve as a soft gate to trigger additional residual learning blocks, and the pose information was not fully used to guide the subsequent frontalization process. Instead, we use the universal pose feature extracted by HopeNet to retain the maximum pose information. As shown in Figure~\ref{fig:pretrain}, two non-linear fully connected layers are employed as pose mapping functions that convert the pose feature to match the same dimension as the profile face feature. Subsequently, the pose feature will serve as soft gates to control the pose frontalization process.

\textbf{Frontalization transformation:} 
The goal of the proposed FSPFM is to learn a mapping function $M_g$ to transform the feature of profile image $I_p$ such that it approximates the feature of frontal image $I_f$, namely $M_g(\phi(I_p)) \approx \phi(I_f)$, where $\phi$ is the face feature extractor. Similar to~\cite{cao_pose_2018, huang_attention_2021}, this work frames the transformation process as learning a residual between the profile and frontal features. Specifically, the mapping function $M_g$ is formulated as two stacked blocks of neural networks that adjust the profile feature through additive residuals to approach its frontal counterpart. 
Moreover, unlike previous explorations that used a single pose label as a soft gate, we leverage the entire pose feature as an element-wise soft gate to better control the learned residual.

The frontalization transformation can be formulated as follows. Given a profile image $I_p$, face feature extractor $\phi(\cdot)$, pose estimator $\theta(\cdot)$, residual blocks $\mathcal{T}_1(\cdot), \mathcal{T}_2(\cdot)$, and pose mapping functions $ \mathcal{P}_1(\cdot), \mathcal{P}_2(\cdot)$, the frontalized feature is obtained by adding the profile feature and the learned residual controlled by the prior pose information, as expressed in the following:
% we intend to obtain the frontalized feature by adding up the profile feature and the learned residual controlled by prior pose information, expressed as follows:
\begin{equation}
\begin{split}
    M_g\phi(I_p) = \phi(I_p) & + \mathcal{T}_1(\phi(I_p)) \odot \mathcal{P}_1(\theta(I_p)) \\
    & + \mathcal{T}_2(\phi(I_p)) \odot \mathcal{P}_2(\theta(I_p)) \\
    \approx \phi(I_f) .
\end{split}
\end{equation}

Previous explorations~\cite{huang_attention_2021} often relied on pre-computed pose labels to control a progressive frontalization process to ensure accuracy. In contrast, our proposed FSPFM architecture dynamically computes pose embeddings for each input and leverages it as an element-wise soft gate to control the learned residual more precisely.
Moreover, FSPFM takes into account the full pose embedding, including yaw, pitch, and roll. This enables a better understanding of pose variations and enhances the performance across a larger range of pose differences.

\subsection{Frontal-profile feature adaptation} 
\label{Section3-2}

This subsection provides a new training paradigm to optimize FSPFM in order to maximize its potential by treating it as a domain adaptation task. Specifically, two face recognition networks are introduced to extract frontal and profile features, respectively, denoted as F-Net and P-Net. The F-Net provides frontal face features as supervision to guide the pose frontalization process in the P-Net and to bridge the domain gap between them. To achieve this, a two-stage training process and a self-attention module are proposed to facilitate the domain adaptation.

\begin{table*}[t]
  \centering
  \caption{Performance of the proposed method on two cross-pose datasets (CPLFW, CPF-FP), one standard dataset (LFW), and two cross-age datasets (CALFW, AgeDB-30). Several notations are explained as follows. FT: enable fine-tuning stage; ADA: attention-guided domain adaptation loss.}
  \begin{adjustbox}{width=0.9\textwidth}
    \begin{tabular}{ccccc|ccccc}
    \toprule
    Backbone & Dataset & FSPFM   & FT    & ADA    & \multicolumn{1}{c}{CPLFW} & \multicolumn{1}{c}{CFP-FP} & \multicolumn{1}{c}{LFW} & \multicolumn{1}{c}{CALFW} & \multicolumn{1}{c}{AgeDB-30} \\
    \midrule
    \multirow{5}[2]{*}{ResNet50} & MS1M &  -    &  -    &  -    & 92.75 & 98.43 & 99.75 & 96.00 & 97.95 \\
          & MS1M+PoseSim &  -    &  -    &  -    & 93.37	& 98.48 & 99.75 & 96.00 & 98.05 \\
          & MS1M+PoseSim & \checkmark   &  -    &  -    & 93.58 & 98.60 & 99.82 & 96.03 & 98.13 \\
          & MS1M+PoseSim & \checkmark   & \checkmark   &  -    & 93.75 & 98.64 & \textbf{99.83} & \textbf{96.05} & 98.05 \\
          & MS1M+PoseSim & \checkmark   & \checkmark   & \checkmark   & \textbf{93.83} & \textbf{98.67} & \textbf{99.83} & \textbf{96.05} & \textbf{98.13} \\
    \bottomrule
    \end{tabular}%
    \end{adjustbox}
  \label{tab:pifr-results}%
\end{table*}%

\textbf{Two-stage training strategy:}
The training process mainly comprises two stages, i.e., pre-training the entire face recognition pipeline and fine-tuning the FSPFM. 

In the first stage, F-Net and P-Net share the same architecture and weights. As shown in Figure~\ref{fig:pretrain}, the ResNet-based feature extractor and the proposed FSPFM are trained from scratch on a large-scale training set. The pre-training step not only allows the feature extractor to learn the capability of general face representation but also FSPFM to obtain proper initial parameters. Then, F-Net and P-Net will inherit the weights from the pre-trained models. 

The second stage involves pair-wise fine-tuning to refine the FSPFM in the P-Net. To begin with, a synthetic dataset~\cite{huang_attention_2021} is selected, which consists of simulated profile images based on the data in the MS1M~\cite{guo_ms-celeb-1m_2016} dataset. Paired training data is constructed by selecting frontal images from the MS1M dataset and corresponding profile images from the MS1M-Pose-Sim dataset, with pose angles ranging from 40$^\circ$ to 90$^\circ$. Subsequently, as illustrated in Figure~\ref{fig:finetune}, the frontal and profile images are fed to the pre-trained F-Net and P-Net respectively to obtain corresponding face features. The profile feature will be fed to the trainable FSPFM to perform feature-space frontalization. Finally, a domain adaptation loss will align the frontal and frontalized features to optimize the FSPFM in P-Net and bridge the domain gap.

\textbf{Attention-guided feature adaptation:} 
The concurrent feature adaptation paradigm mitigates the domain gap between frontal and profile features by minimizing the domain adaptation loss in the feature space. However, the pose difference between a frontal and profile image is not necessarily distributed equally across all feature dimensions. Existing studies in explainable face recognition methods~\cite{xu_discriminative_2023, lu_explainable_2024} have revealed that each dimension of an extracted face feature corresponds to specific regions in the face image. Therefore, the most discriminative differences between the frontal and profile faces caused by pose variations often reflect on specific feature dimensions. The adaptation process can be further enhanced by identifying those pose-sensitive feature dimensions. 
Prior work~\cite{huang_attention_2021} employed a similar idea of re-weighting the feature adaptation process, by simply selecting the maximum dimension of the frontal feature for loss calculation. 
However, the feature dimension with maximum value is not necessarily the most discriminative for face recognition. 
Instead, we introduce an attention module to F-Net in the fine-tuning stage, which automatically re-weights the frontal feature in a self-adaptive manner. Specifically, it takes extracted frontal features as input and generates a weight vector that learns to identify the most pose-discriminative feature dimensions during training. This attention module in F-Net will be jointly optimized with the FSPFM in the P-Net under the supervision of the attention-guided domain adaptation loss.

\subsection{Loss functions}
This work employs the ArcFace loss $\mathcal{L}_{arcface}$~\cite{deng_arcface_2019} in both pre-training and fine-tuning stages to learn the general discriminative face features. Then, the proposed attention-guided domain adaptation loss $\mathcal{L}_{ada}$ is introduced to the fine-tuning stage to supervise the profile feature frontalization:
\begin{align}
\mathcal{L}_{ada} = \frac{1}{N} \sum_{i=1}^N \lVert M_g \phi(I^i_f) \odot AM(\phi(I^i_f)) - M_g \phi(I^i_p)\rVert^2_2 ,
\end{align}
where $I_f$ and $I_p$ refer to the frontal and profile face images, $\phi(\cdot)$ is the feature extractor, $M_g$ refers to FSPFM, $AM(\cdot)$ is the attention layer, and $N$ denotes the batch size. 
% the face frontalization transformation mapping

The overall loss function is a weighted combination of the recognition loss and the attention-guided domain adaptation loss:
\begin{align}
\mathcal{L}_{total} &= \mathcal{L}_{arcface} + \lambda \mathcal{L}_{ada},
\end{align}
where the weight $\lambda$ is set as 4 in the fine-tuning stage.

\section{Experimental Results}
\subsection{Experimental setups}

\subsubsection{Datasets}

The cleaned version of the MS1M~\cite{guo_ms-celeb-1m_2016} dataset released by Deng et al.~\cite{deng_arcface_2019}\footnote{MS1M-RetinaFace in \url{https://github.com/deepinsight/insightface/tree/master/recognition/_datasets_}} is used as the training set, which comprises 5.1M face images belonging to 93k identities. 
We also leverage MS1M-Pose-Sim~\cite{huang_attention_2021}, the synthetic dataset with pose-variant face images for pair-wise fine-tuning. 
This dataset contains 532,244 images with pose angles ranging from 0$^\circ$ to 90$^\circ$. The entire dataset is used in the pre-training stage as augmented data, while only those with more than 40$^\circ$ yaw, pitch, or roll angles are used for the fine-tuning stage. 
Additionally, the original MS1M-Pose-Sim dataset provided by~\cite{huang_attention_2021} was not properly preprocessed. This work leverages RetinaFace~\cite{deng_retinaface_2020} for face detection and crops the simulated images in the same format as those from MS1M. 
During the evaluation, two cross-pose datasets are employed, namely CPLFW~\cite{zheng_cross-pose_2018} and CFP-FP~\cite{Sengupta_frontal_2016}. In addition, three other popular datasets are used, namely LFW~\cite{huang_labeled_2008}, CALFW~\cite{zheng_cross-age_2017}, and AgeDB-30~\cite{moschoglou_agedb_2017}. %, and IJB-C~\cite{maze_iarpa_2018}. 

\subsubsection{Implementation details}
The ResNet50~\cite{he_deep_2016} backbone feature extractor and ArcFace loss~\cite{deng_arcface_2019} have been adopted throughout all experiments. As elaborated in Section~\ref{Section3-2}, the training process contains two stages, i.e., pre-training and fine-tuning. In the pre-training stage, the face recognition model is trained on a mixture of the MS1M dataset and the simulated dataset from scratch. It is trained for 20 epochs using the SGD optimizer with a batch size of 256. The learning rate is set to 0.1 and is decreased by a factor of 0.1 at epochs 10, 13, 16, and 18. The weights of the pre-trained model are inherited by both F-Net and P-Net for the second stage of training. In the fine-tuning stage, the feature extractors in both F-Net and P-Net remain frozen, while the attention layer in F-Net and the FSPFM in P-Net will be jointly optimized. The learning rate is set to 0.001 and they are fine-tuned for 50 epochs.
Throughout all experiments, the random seed is selected as 42 to ensure reproduction, and the following data augmentation operations are used: random grayscaling, random horizontal flipping, and color jittering. 
% , momentum of 0.9, and weight decay of 0.0001

\subsection{Experimental results}
\subsubsection{Performance on multiple datasets}
Table~\ref{tab:pifr-results} presents verification results of the proposed method on two widely recognized cross-pose datasets, CPLFW and CFP-FP. The results systematically showcase the effectiveness of each proposed module in the PIFR task. Furthermore, the proposed method has been assessed on three additional face recognition benchmarks, further demonstrating its consistent high accuracy across a diverse range of face recognition scenarios.

The first two rows show that simply augmenting the MS1M dataset with the synthetic MS1M-Pose-Sim dataset marginally improves the performance on two cross-pose datasets CPLFW and CFP-FP, as well as on the AgeDB-30 dataset. Training the model with the proposed FSPFM from scratch on the augmented dataset improves recognition performance across all five datasets relative to the baseline model, particularly on the two cross-pose datasets. Fine-tuning the model with domain adaptation loss further enhances the performance on CPLFW and CFP-FP datasets compared to its counterpart trained from scratch. Consequently, the attention module boosts the performance across all five datasets, with a particularly notable improvement of 1.08\% on CPLFW relative to the baseline.

\begin{table}[t]
  \centering
  \caption{Performance comparison among state-of-the-art PIFR methods on two cross-pose datasets (CPLFW, CPF-FP), one standard dataset (LFW), and two cross-age datasets (CALFW, AgeDB-30).}
  \begin{adjustbox}{width=\linewidth}
    \begin{tabular}{c|ccccc}
    \toprule
    Method & CPLFW & CFP-FP & LFW   & CALFW & AgeDB-30 \\
    \midrule
    DREAM$^\dagger$ ~\cite{cao_pose_2018} & 92.65 & 97.60 & 99.82 & 96.08 & 98.05 \\
    PAM~\cite{tsai_pam_2021}   &  92.80 & 97.89 & 99.82 & 96.05 & 98.00 \\
    PAD~\cite{zhang_pose_2022}   &  -    & 97.78 & 99.77 &  -    &  -  \\
    PCTN~\cite{zeng_implicit_2023} & 92.97 & - & 99.82 & \textbf{96.13} & - \\ 
    AGPM$^{\dagger\dagger}$~\cite{huang_attention_2021}  &  91.23 & 96.93 &  99.60 & 94.57  & 95.00 \\
    \textbf{Ours}  &  \textbf{93.83} & \textbf{98.67} & \textbf{99.83} & 96.05 & \textbf{98.13} \\
    \bottomrule
    \multicolumn{6}{l}{\footnotesize $\dagger$: based on reproduction of PAM authors; $\dagger\dagger$: based on our reproduction.} \\

    \end{tabular}%
    \end{adjustbox}
  \label{tab:pifr-compare}%
\end{table}%

\begin{table}[t]
  \centering
  \caption{Performance comparison between different attention-guided domain adaptation (ADA) losses.} %and a pipeline without ADA.
  \begin{adjustbox}{width=\linewidth}
    \begin{tabular}{c|ccccc}
    \toprule
          & \multicolumn{1}{c}{CPLFW} & \multicolumn{1}{c}{CFP-FP} & \multicolumn{1}{c}{LFW} & \multicolumn{1}{c}{CALFW} & \multicolumn{1}{c}{AgeDB-30}  \\
    \midrule
    No ADA & 93.75 & 98.64 & 99.83 & 96.05 & 98.05  \\
    Huang's ADA~\cite{huang_attention_2021} & 93.80  & 98.61 & 99.78 & \textbf{96.10}  & 98.10  \\
    Proposed ADA & \textbf{93.83} & \textbf{98.67} & \textbf{99.83} & 96.05 & \textbf{98.13} \\
    \bottomrule
    \end{tabular}%
    \end{adjustbox}
  \label{tab:ablation}%
\end{table}%

\subsubsection{Comparison to state-of-the-art PIFR methods}

Table~\ref{tab:pifr-compare} shows the comparison of the proposed method to several state-of-the-art approaches, namely DREAM~\cite{cao_pose_2018}, PAM~\cite{tsai_pam_2021}, PAD~\cite{zhang_pose_2022}, PCTN~\cite{zeng_implicit_2023}, and AGPM~\cite{huang_attention_2021}. To ensure a fair comparison, all the selected methods employ ResNet50 as a feature extractor. We re-trained the AGPM method with this backbone for better results. Due to a lack of training code, the highest-performing scores for DREAM, PAM, PAD, and PCTN are directly taken from their publications. 
The results show that our method obtains better scores on most datasets. In particular, it significantly outperforms the current state-of-the-art approaches in two cross-pose scenarios, i.e., CPLFW and CFP-FP, and also maintains superior results on LFW and AgeDB-30. On the other hand, PCTN reports the highest score on CALFW among the compared methods. Despite a similar mechanism of feature-space pose transformation, DREAM and AGPM show relatively lower accuracy on the two cross-pose datasets, showcasing the advantage of the proposed method. 

\subsubsection{Effectiveness of attention-guided feature adaptation loss}

This subsection further validates the effectiveness of the proposed attention-guided feature adaptation loss. 
For a fair comparison, we train the same PIFR pipeline on the same dataset using our proposed attention-guided feature adaptation loss and the loss proposed by~\cite{huang_attention_2021}, respectively. 

The results in Table~\ref{tab:ablation}  show that the idea of re-weighting the feature adaptation process improves overall recognition performance compared to the baseline.
The attention loss proposed by~\cite{huang_attention_2021} increases accuracy on some datasets while sacrificing performance on CFP-FP and LFW. Moreover, our proposed attention loss achieves better improvement in most scenarios, in particular on the two cross-pose datasets.

\section{Conclusion}
This paper addressed the critical problem of PIFR. A novel pose frontalization module, called FSPFM, was first proposed to bridge the gap between frontal and profile images in the feature space.  
Subsequently, a two-stage training paradigm was proposed to maximize the effectiveness of FSPFM. It consisted of an initial pre-training stage to learn general face representation, followed by a fine-tuning stage supervised by a novel attention-guided domain adaptation loss.
The evaluation results show that the proposed method significantly enhances the performance on multiple benchmarks, in particular on two cross-pose datasets, and outperforms current state-of-the-art approaches. The results highlight the potential of feature-space frontalization, offering an effective and robust solution for real-world applications of face recognition under varying poses.


% \let\oldbibliography\thebibliography
% \renewcommand{\thebibliography}[1]{%
%   \oldbibliography{#1}%
%   \setlength{\itemsep}{-1pt}%
% }

\bibliographystyle{IEEEbib}
{\footnotesize
\bibliography{refs}}

\end{document}